\documentclass[12pt]{article}

\usepackage{graphicx}
\usepackage{setspace}
\usepackage{color}
\usepackage{booktabs}
\usepackage{colortbl}

\newcommand{\beq}{\begin{equation}}
\newcommand{\eeq}{\end{equation}}
\newcommand{\beqs}{\begin{eqnarray}}
\newcommand{\eeqs}{\end{eqnarray}}

\newcommand{\vdelta}{\vec{\delta}}
\newcommand{\vx}{{\bf x}}

\newcommand{\lsim}{\:\raise -4pt\hbox{$\stackrel{\textstyle <} {\sim}$}\:}
\newcommand{\prob}{{\rm Prob}}
\newcommand{\sigrr}{{\hat\sigma_r}}
\newcommand{\ignore}[1]{}
\begin{document}

\bibliographystyle{plain}
%

\begin{titlepage}
\small
\begin{flushright}
LU TP 94-19\\
October 1994
\end{flushright}
\normalsize
\vspace{0.1in}
\large
\begin{center}
{\bf Estimating nonlinear regression errors without doing regression}\footnote{This note contains derivations of the formalism and elaborations of the results presented in C. Peterson, "Determining dependency structures and estimating nonlinear regression errors without doing regression", {\it International Journal of Modern Physics} {\bf 6}, 611-616 (1995).}\\
\vspace{0.3in}
\normalsize
Hong Pi \\
\medskip	
Department of Computer Science \& Engineering, \\ 
Oregon Graduate Institute, P.O. Box 91000, Portland, Oregon 97291-1000\\
\medskip
\medskip
Carsten Peterson \\
\medskip	
Department of Theoretical Physics, University of Lund,\\
S\"{o}lvegatan 14A, S-223 62, Lund, Sweden\\
Email: carsten@thep.lu.se

\vspace{0.1in}

\end{center}
\vspace{0.2in}

\normalsize
{\bf Abstract:}

A method for estimating nonlinear regression errors and their distributions
without performing regression is presented.  Assuming continuity of the 
modeling function the variance is given in terms of conditional probabilities 
extracted from the data.  For $N$ data points the computational demand
is  $N^2$. Comparing the predicted residual errors with those derived from 
a linear model assumption provides a signal for nonlinearity. The method is 
successfully illustrated with data generated by the Ikeda and Lorenz maps 
augmented with noise.  As a by-product the embedding dimensions of these 
maps are also extracted.

  
\end{titlepage}
\normalsize

\subsection*{Background}

Most measurements of physical processes are noisy. This is often due to 
the fact that all independent variables have not been measured. Being able to 
estimate the noise distribution and its variance directly from data with no 
assumptions about the underlying signal function is most desirable.
 It would provide a natural step prior to any modeling of a system 
 (e.g. artificial neural network) since one then knows the optimal performance 
limit of the fit in advance. Furthermore, methods for filtering data often 
require prior estimate of noise variance. 

To be more concrete,  given a table of data  $\{(y^{(i)}, \vx^{(i)})$, $i = 1,2,...,N\}$,  
where $y$ is the dependent variable and the $d$-dimensional vector $\vx$ denotes 
the set of explanatory variables, one aims at estimating the variance of 
$r$ ($\sigma_r^2$) for  
\beq
\label{fx}
 \hat{y} = F(\vx) + r
\label{F}
\eeq 
where $F$ represents the optimum model.

Conventional procedures for estimating $\sigma_r^2$ are model-based. 
One fits the data to a model, a particular choice of $F$, and then interprets 
the deviation of the fit as noise. In the special case of 
linear regression models \cite{ham} where $F$ takes the form
\beq
 \hat{y} = a_0 + \sum^d_{k=1} a_k x_k,
\label{lin}
\eeq 
a sample estimate for $\sigma_r^2$ is explicitly given by 
\beq
\sigma_r^2 =  \sigma^2 - \sum^d_{k=1} a_k \langle y, x_k \rangle 
\label{sigma_lin}
\eeq 
where $\sigma $ denotes the $y$-variable variance and the angled brackets  
covariances. In this letter we devise a method for estimating the optimum $\sigma_r$ 
when the modeling function $F$ is not restricted to be linear.  The
estimate does not rely on any conjecture about the form of $F$.  
The only assumption is that $F$ is uniformly 
continuous. Loosely speaking we extract probability densities from the data 
and by sampling data with decreasing bin sizes such that any noise point will appear 
as a discontinuity. The concept of using the requirement of continuity for establishing 
dependencies on $x_k$ was previously explored in the $\delta$-test method 
\cite{delta}, where noise levels had to be estimated by making assumptions
about the probability distributions of $r$. 

The approach in this work is novel and unique as compared to other methods and 
also with respect to ref. \cite{delta} since  no assumption about the distribution 
of $r$ is needed -- $\sigma_r^2$ is computed directly as an integral 
over data densities. Actually,  the method also disentangles different noise distributions.

Comparing the obtained $\sigma_r^2$  with what is extracted assuming 
a linear model in Eq. (\ref{sigma_lin}) provides means for establishing 
nonlinearities.

We illustrate the power of the method with two examples of chaotic time series 
augmented with noise: the Ikeda \cite{ikeda} and Lorenz \cite{lor} maps.
In addition to finding noise levels, the method can also be used for   
determining embedding dimensions.

\subsection*{Method}

The goal is to derive a statistical estimate on the variance of $r$ for 
the optimum 
model describing Eq.~(\ref{F}). By optimum we mean a model $F$ (Eq. (\ref{fx})) 
such that $r$ and $\hat y$ are uncorrelated and that $r$ represents identically 
and independently distributed ({\bf i.i.d}) noise.  Most adaptive algorithms such 
as neural network models are designed to find such an optimum function.  

We use the conditional probabilities defined in the $\delta$-test 
\cite{delta} -- for a pair of positive real numbers $\epsilon$ and
$\delta$, one constructs directly from the data the conditional probability 
\beq
\label{pepsdelta}
P(\epsilon|\;\delta) \equiv
 P(|\Delta y| \leq \epsilon \;|\; |\Delta\vx| \leq\delta)
\eeq 
where $|\Delta\vx| \equiv \max_k |x_k - x^\prime_k|$.
In the limit $\delta \rightarrow 0$, one obtains
\begin{eqnarray}
\label{peps}
P(\epsilon) &\equiv& \lim_{\delta\rightarrow0} P(\epsilon|\;\delta) \nonumber\\
 &=& P(|F(\vx) - F(\vx^\prime) + r - r^\prime| \leq \epsilon \;|\;  
  |\vx-\vx^\prime|\rightarrow 0)  \nonumber\\
 &=& \prob(|\Delta r| \leq \epsilon), 
\end{eqnarray}
where the property of function continuity,   
$F(\vx)-F(\vx^\prime)\rightarrow 0$ for $\vx\rightarrow \vx^\prime$,  
is exploited. Eq.~(\ref{peps}) establishes a 
connection between the probability distribution of the residuals 
$\rho(|\Delta r|)$ 
to the quantity $P(\epsilon)$,
 which is directly calculable from the data sample.

The probability density of the residual separation $|\Delta r|$
is given by
\beqs
\label{rhor}
      \rho(|\Delta r|) &=& 
      -\frac{d}{d|\Delta r|} \prob(|\Delta r^\prime| > |\Delta r|) \nonumber\\
     &=& \left[\frac{d}{d\epsilon} P(\epsilon)\right]_{\epsilon=|\Delta r|}.
\eeqs
Thus  moments of $|\Delta r|$  can be related to $P(\epsilon)$  using eqs.~(\ref{peps},  \ref{rhor}). With partial integration one obtains
\beq
\label{deltarn}
     \langle |\Delta r|^n \rangle
 = n \int_0^\infty d\epsilon\;\epsilon^{n-1}\, [ 1 - P(\epsilon)]
\eeq
If $r$ is {\bf i.i.d} one has $\langle (\Delta r)^2\rangle = 2\sigma_r^2$.
Our estimate for the residual variance of the optimum model is then given by
\beq
\label{rvar}
\sigma_r^2 = \int_0^\infty d \epsilon\; \epsilon\, [1-P(\epsilon)].
\eeq
We note that the integrand in Eq. (\ref{rvar}) suppresses the small $\epsilon$ region. 
This feature is desirable in limited statistics situations with few  high 
resolution (small $\epsilon$) data points.  For higher moments this effect is even 
further pronounced.  In addition to the variance,  Eq. (\ref{deltarn}) of course also 
provides us with the skewness of the distribution (n=3).  

Eq.~(\ref{rhor}) shows that $P(\epsilon)$ measures the cumulative 
distribution of the noise separations.  It is then possible to discern
the noise distribution through $P(\epsilon)$. 

In some sense Eq.~(\ref{deltarn}) is equivalent to calculating the expectation 
value \cite{rip},
\beq
\label{sigexpr}
  \sigma_r^2 = E[ (\Delta y)^2 |\, |\Delta\vx| \leq \delta ]_{\delta\rightarrow0}.
\eeq
which may  be easier to implement numerically,  if one only wants to estimate 
$\sigma_r^2$.

\begin{figure}[hb]
\begin{center}
\includegraphics[width=0.5\columnwidth, angle=90]{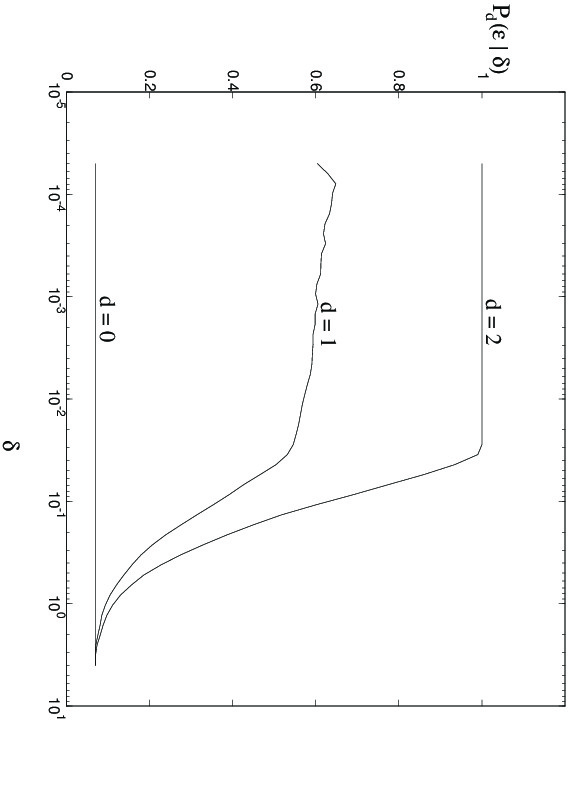}
\caption{\label{ped}$P_{d}(\epsilon\vert\,\vdelta)$ as a function of
 $\delta$ at a fixed $\epsilon = 0.108$ for the H\'enon map  
 [$x_{t} = 1 - 1.4x^2_{t-1} + 0.3 x_{t-2}$] with N=3000 data points.}
\end{center}
\end{figure}

So far we have assumed an infinite amount of data. Some implementation issues 
are pertinent with limited statistics situations.
A suitable binning of the $\log\epsilon$-$\log\delta$ plane for 
evaluating the probabilities and estimating their statistical errors 
is given in \cite{delta}.  In Fig.~\ref{ped} a typical example of 
$P(\epsilon|\;\delta)$ is shown. For a fixed  $\epsilon$, $P(\epsilon|\;\delta)$ 
rises and reaches a plateau  as $\delta $ decreases.  $P(\epsilon)$ is determined by
the largest point with acceptable statistics in the plateau region.
\beq
\label{pdmax}
 P(\epsilon) \equiv \max_{\delta>0} P(\epsilon\vert\,\delta)
\eeq
The integral of Eq.~(\ref{rvar}) is easily computed with e.g. the Simpson method. 

%
%

\subsection*{Explorations}

{\bf The Ikeda Map}. This system \cite{ikeda} describes the evolution of a laser in a ring cavity 
with a lossy active medium. In terms of the complex variable 
$z_{t} = x_{t} + i\,y_{t}$, the map is defined by
\beq
\label{ikemap}
  z_{t+1} = p + B\, z_{t}\exp[i\kappa - \frac{i\alpha}{1+|z_{t}|^2}].
\eeq
Sets of $N=2000$ data points are generated using Eq.~(\ref{ikemap})
 with the parameters $p=1.0$,  $B=0.9$, $\kappa=0.4$ and $\alpha=6.0$  \cite{brown}, 
and with Gaussian noise 
added to the $x$ component at the each iteration as 
$x_{t} = x_{t} + r$ with standard deviations 
$\sigma_r$=0.0, 0.01, 0.02 and 0.03 respectively.  
%




\begin{figure}[ht]
\begin{center}
\includegraphics[width=0.5\columnwidth, angle=90]{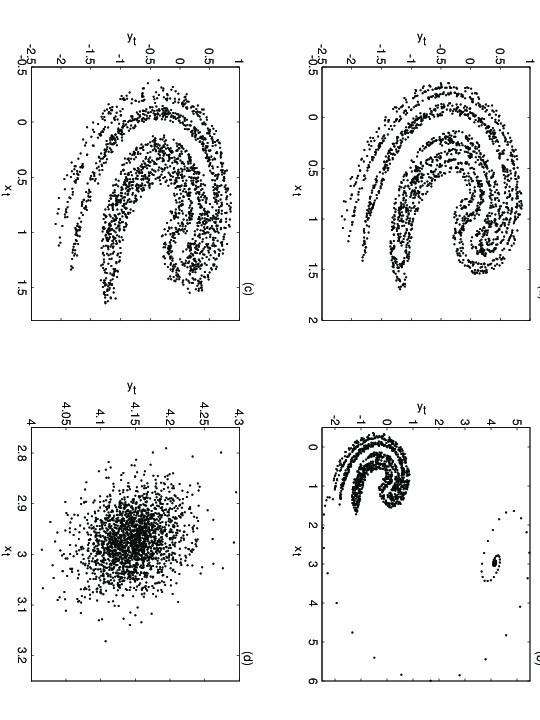}
\caption{\label{ikeda}The Ikeda map shown in its $x$-$y$ phase space.  A Gaussian noise term 
 with standard deviation $\sigma_r$ = 0.0 (a), 0.01 (b), 0.02 (c)
 and 0.03 (d) is added iteratively to the $x_{t}$-component.}
\end{center}
\end{figure}

We now apply our method to estimate what the error would have been if we had 
regressed $x_{t}$ on various sets of explanatory variables. The results are shown in 
\begin{table}[h,t,b]
\begin{center}
\begin{tabular}{|c||c|c||c|c||c|c||c|c|}   \hline\hline
   $\sigma_r$ & \multicolumn{2}{c||}{0.00}
                      & \multicolumn{2}{c||}{0.01}
                      & \multicolumn{2}{c||}{0.02}
                      & \multicolumn{2}{c|}{0.03}  \\ \hline
$\sigma_r/\sigma$   & \multicolumn{2}{c||}{0.0000}
                                 & \multicolumn{2}{c||}{0.0208}
                                 & \multicolumn{2}{c||}{0.0424}
                                 & \multicolumn{2}{c|}{0.5621}  \\ \hline
Variables
           & $\left(\sigrr\right)_{LR}$ & $\left(\sigrr\right)_{NL}$ 
           & $\left(\sigrr\right)_{LR}$ & $\left(\sigrr\right)_{NL}$ 
           & $\left(\sigrr\right)_{LR}$ & $\left(\sigrr\right)_{NL}$ 
           & $\left(\sigrr\right)_{LR}$ & $\left(\sigrr\right)_{NL}$ 
          \\ \hline
\{none\} & 1.000 & 1.003 & 1.000 & 1.003 & 1.000 & 1.003 & 1.000 & 1.001
       \\ \hline
\{$x_{t-1}$\} & 0.997 & 0.819 & 0.998 & 0.792 & 0.994 & 0.813 
        & 0.642 & 0.644       \\ \hline
\{$x_{t-1}$, $y_{t-1}$\} & 0.887 & 0.0055 & 0.889 & 0.021 & 0.884 & 0.044
        & 0.557 & 0.563       \\ \hline\hline
\end{tabular}
\end{center}
\caption{Regression errors on $x_{t}$ expressed as fractional errors
 $\sigrr$ for 
various sets of explanatory 
variables.  The subscripts {\bf LR} and {\bf NL} stand for linear regression 
(Eq. (\protect\ref{sigma_lin})) and our method allowing for nonlinear 
dependencies (Eq.~(\protect\ref{rvar})), respectively. Due to the effect of the noise, the noise fraction $\sigma_r/\sigma $� varies considerably for differing noise levels.}
\end{table}
Table 1, which shows that one needs to use the explanatory variable
set $\{x_{t-1}$, $y_{t-1}\}$ in order to reduce the residual variance
to the optimum level, and our method 
gives quite accurate estimates on that level 
in terms of the noise fraction $\sigma_r/\sigma$.
In the case of $\sigma_r = 0.0$,
the linear regression model gives a noise level 0.887, while our
method identifies a negligible noise level ($0.005$).  This indicates
that the dependency of $x_t$ on $x_{t-1}$ and $y_{t-1}$ is predominantly
nonlinear. Such a signature of nonlinearity exists as long as the noise
level is modest -- below $\sigma_r=0.02$ in this case.  This is
consistent with what can be seen in Fig.~\ref{ikeda}, where the nonlinear
structure clearly disappears in (d) when the noise reaches $\sigma_r = 0.03$. 

\begin{figure}[ht]
\begin{center}
\includegraphics[width=0.5\columnwidth, angle=90]{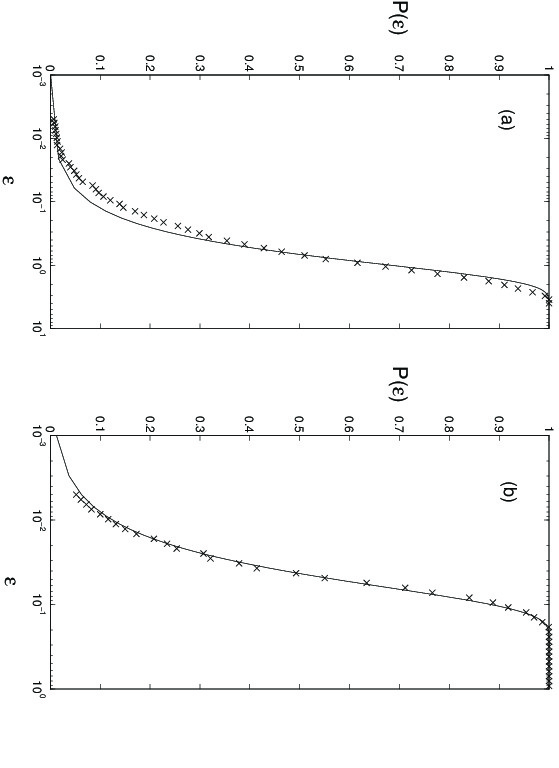}
\caption{\label{ikedapeps}$P(\epsilon)$ versus $\epsilon$ for the Ikeda map with Gaussian
 fractional noise $\sigma_r/\sigma = 0.0424$.  The symbols are
 the values calculated from the data.  
 (a). $P(\epsilon)$ based on using $x_{t-1}$ as the explanatory variable. 
 The line is the analytic expression in Eq.~(\protect\ref{errorf}) with $\sigma_r = 0.71$.
 (b). $P(\epsilon)$ based on using $\{x_{t-1}, y_{t-1}\}$ as the explanatory 
 variables. 
 The line is the analytic expression in Eq.~(\protect\ref{errorf}) with $\sigma_r = 0.045$.}
\end{center}
\end{figure}

Next we compare  $P(\epsilon)$ calculated with our method from data with 
what is expected from a  Gaussian distribution with standard deviation $\sigma_r$ 
\beq
\label{errorf}
P(\epsilon) = \int_0^\epsilon \rho(|\Delta r|)\,d|\Delta r| = erf(\frac{\epsilon}{2\sigma_r}),
\eeq
where $erf$(.) is the error function.  In Fig.~3  $P(\epsilon)$  is shown together 
with the Gaussian analytic expression in Eq.~(\ref{errorf}).  The lines  correspond 
to a least-mean-square fit with $\sigma_r$ as parameter.  The misfit in 
Fig.~\ref{ikedapeps}a indicates that the residuals based on $x_t = F(x_{t-1}) + r$ would be non-Gaussian 
distributed, and that more explanatory 
variables may be needed to model the process.  Fig.~\ref{ikedapeps}b shows 
the $P(\epsilon)$ based on using $x_{t-1}$ and $y_{t-1}$ as the explanatory 
variables.  It indicates that the residuals can be reduced to a Gaussian process 
if $x_t$ is nonlinearly regressed on $x_{t-1}$ and $y_{t-1}$.  The best fit 
results in $\hat\sigma_r = 0.045$ in agreement with the estimate given in Table 1. 
The absence of sufficient explanatory variables in  Fig.~\ref{ikedapeps}a emulates 
additional noise, which also would manifest itself in a skew distribution with a 
nonvanishing $\langle |\Delta r|^3\rangle$.

Let us next turn to the problem of determining the embedding dimension within this 
scheme assuming that the only data we have at our disposal are the observations of 
the $x$ component. The  variance estimate can then be used to identify the 
minimum embedding dimension, in a procedure similar to the $\delta$-test \cite{delta}.  
 What we need to do is to find the (smallest) set of variables that minimizes the 
residual error. The results are given in Table 2,
\begin{table}
\begin{center}
\begin{tabular}{|c||c|c||c|c||c|c|}   \hline\hline
   $\sigma_r$   & \multicolumn{2}{c||}{0.00}
                      & \multicolumn{2}{c||}{0.02}
                      & \multicolumn{2}{c|}{0.03}  \\ \hline
         $\sigma_r /\sigma$       & \multicolumn{2}{c||}{0.0000}
                                                & \multicolumn{2}{c||}{0.0424}
                                                 & \multicolumn{2}{c|}{0.5621}  \\ \hline \hline
Variables 
           & $\left(\sigrr\right)_{LR}$ & $\left(\sigrr\right)_{NL}$ 
           & $\left(\sigrr\right)_{LR}$ & $\left(\sigrr\right)_{NL}$ 
           & $\left(\sigrr\right)_{LR}$ & $\left(\sigrr\right)_{NL}$ 
          \\ \hline \hline
\{none\} & 1.000 & 1.003 & 1.000 & 1.003 & 1.000 & 1.001
       \\ \hline
\{$x_{t-1}$\} & 0.997 & 0.819  & 0.994 & 0.813 
        & 0.642 & 0.644       \\ \hline
\{$x_{t-1},\;x_{t-2}$\} & 0.954 & 0.505 & 0.950 & 0.623
        & 0.622 & 0.636       \\ \hline\hline
\{$x_{t-k}\; k=1,..3$\} & 0.946 & 0.077 & 0.946 & 0.195
        & 0.611 & 0.635       \\ \hline\hline
\{$x_{t-k}\;k=1,..4$\} & 0.936 & 0.025 & 0.931 & 0.074
        & 0.603 & 0.636       \\ \hline\hline
\{$x_{t-k}\; k=1,..5$\} & 0.934 & 0.026 & 0.931 & 0.077
        & 0.598 & 0.645       \\ \hline\hline
\end{tabular}
\end{center}
\caption{Regression error on $x_{t}$ for various sets of explanatory 
variables. Same notation as in Table 1.}
\end{table}
from which we see that the residual error ceases to  decrease
at $k=4$.  Therefore we identify the embedding dimension $d_E = 5$
for the Ikeda map, as long as the noise level is not too high ($\sigma_r<$ 0.03). 
Using the {\it False Nearest Neighbors} method, one finds $d_E = 4$ \cite{brown}.
As shown in the Table 2,  $d_E = 4$ would result in quite small a 
residual error of $0.077$ and therefore provides a fairly good
embedding. However, $d_E = 5$ is a better choice.

We observe that in the results presented above, the error estimate
$\sigma_r$ is very close to  $1.0$ when the explanatory variable
set is $\{none\}$, as it should be.  This provides a consistency 
check on the applicability of the method on the particular data set.  
When a significant deviation from $1$ is observed, it indicates that 
some of the assumptions of the method,  e.g.  stationarity, is violated in the data set.
Also note that the noise estimates $\hat\sigma_r$ in Table 2 does not match the 
applied relative noise $\sigma_r/\sigma$ exactly, since the regression 
equation is different from the actual generating process.

\begin{figure}[ht]
\begin{center}
\includegraphics[width=0.7\columnwidth, angle=90]{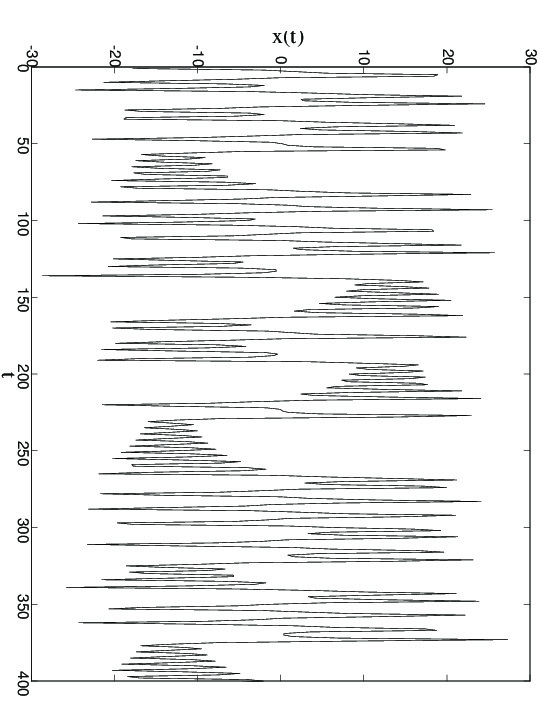}
\caption{\label{lorenxt}The residual variances versus the number of time-lagged variables
 for the Lorenz-$x_{t}$ data.  
 }
\end{center}
\end{figure}

{\bf The Lorenz Map}. A system described by the Lorenz equations \cite{lor}, which describe
meteorological physics,
\beqs
\label{Lorenz}
 \frac{dx_{t}}{dt} &=& \sigma[ - x_{t} + y_{t} ] \nonumber\\
 \frac{dy_{t}}{dt} &=&  r x_{t} - y_{t} -x_{t} z_{t} \\
 \frac{dz_{t}}{dt} &=&  x_{t} y_{t} - b z_{t} \nonumber
\eeqs
can display low-dimensional chaotic behavior when
the parameters are chosen from the chaotic regime \cite{brown}.
We adopt the parameters $r=45.92$, $b=4.0$ and $\sigma=16.0$  \cite{brown}, 
and solve the equations with $\Delta t=0.1$ using the fourth and fifth order
Runge-Kutta methods.
The method is applied iteratively such that the
solutions at $t$ are used as the initial values to the differential
equations to obtain the values at $t+0.1$.  A portion of the data set
is shown in Fig.~\ref{lorenxt}.

%
%

Gaussian noise of various variances are superimposed onto the original 
clean data set.  The results of the variance estimates are shown in
Table~3.  Nonlinearities are evident from the significant differences
between the linear regression errors and the nonlinear estimates.
Based on the values of $\left(\sigrr\right)_{NL}$ we conclude
that three time lag variables are needed to map the variable $x_{t}$.
Hence one has $d_E=4$ for the Lorenz map.

%
\begin{table}[h,t,b]
\begin{center}
\begin{tabular}{|c||c|c||c|c||c|c|}   \hline\hline
   $\sigma_r$ & \multicolumn{2}{c||}{0.00}
                      & \multicolumn{2}{c||}{0.5}
                      & \multicolumn{2}{c|}{1.0}  \\ \hline
$\sigma_r/\sigma$   & \multicolumn{2}{c||}{0.0000}
                                 & \multicolumn{2}{c||}{0.0016}
                                 & \multicolumn{2}{c|}{0.0062}  \\ \hline
           & $\left(\sigrr\right)_{LR}$ & $\left(\sigrr\right)_{NL}$ 
           & $\left(\sigrr\right)_{LR}$ & $\left(\sigrr\right)_{NL}$ 
           & $\left(\sigrr\right)_{LR}$ & $\left(\sigrr\right)_{NL}$ 
          \\ \hline
$k=0$ & 1.000 & 1.000  & 1.000 & 1.000
       & 1.000 & 1.000
       \\ \hline
$k=1$ & 0.670  & 0.520  & 0.691 & 0.517 
        & 0.693 & 0.544       \\ \hline
$k=2$ & 0.653  & 0.084 & 0.653 & 0.110
        & 0.657  & 0.197       \\ \hline\hline
$k=3$ & 0.640  & 0.01  & 0.642 & 0.084
        & 0.646 & 0.152       \\ \hline\hline
$k=4$ & 0.640  & 0.008   & 0.641 & 0.084
        & 0.646 & 0.158       \\ \hline\hline
$k=5$ & 0.634  & 0.008  & 0.635 & 0.084
        & 0.640 & 0.141       \\ \hline\hline
\end{tabular}
\end{center}
\caption{Regression errors on $x_{t}$ expressed as fractional errors
 $\sigrr$ for 
various numbers of time lag variables for the Lorenz map. 
$\sigma_{LR}$ gives the linear regression residual error. 
$\sigma_{NL}$ is the nonlinear
estimate from Eq.~(\protect\ref{sigexpr}).
}
\end{table}

The Gaussian noise we imposed have variances of $0.0$, $0.5$, and $1.0$,
corresponding to the fractional variances of $0.0$, $0.0016$, and
$0.0062$ respectively, which are significantly lower than the 
estimated fractional residual variances, while in the Ikeda-Map example
we had a good match between the true variances and the estimates.
This is not an inconsistency
since for the Ikeda Map, we applied the noise iteratively:
\beq
\label{Frik}
  x_{t+1} = F(x_{t-1}, y_{t-1}) + r,
\eeq
while in the current example 
the noise is superimposed to the signal after the entire sequence of the
signal is generated.
In this case we are dealing
with noisy inputs.  The equation becomes
\beqs
\label{Fr}
  x_{t+1} &=& F[x_{t}-r_{t}, x_{t-1}-r_{t-1}, ..] + r_{t+1} \nonumber \\
	&=& F^\prime[x_{t}, x_{t-1}, ..] + r^\prime_{t+1}.
\eeqs
What the method yields is the variance of the {\bf effective noise} 
$r^\prime$,   
which can be quite different from 
the variance of the superimposed noise $r$.


\subsection*{Summary}

We have developed a general method that efficiently extracts noise variances 
from raw data with no assumptions about the noise distributions. The method 
handles nonlinear dependencies provided that the underlying function is uniformly 
continuous and the noise is additive. The method is not limited to 
determining variances. Any moment of the distribution including skewness and 
also cumulative distributions can  be extracted.

By comparing the extracted noise variances with those derived from 
assumed linear dependencies, signals of nonlinearities are obtained. 

Estimating the variance is very useful for model selection.  
As a by-product the embedding dimensions are obtained in a way 
slightly different from that of ref. \cite{delta}.

We have illustrated the method with two time series examples. The 
method  of course also works in cases with ``horizontal dependencies'' 
-- variables measured at equal times.

Existing approaches to determine dependencies aimed beyond the linear 
regime are either based on entropy measures \cite{kol, fraser} or on 
elaborate autocorrelation measures \cite{grass,brock,savit}. The 
{\it Mutual Information} approach \cite{fraser} has the shortcoming  
that it does not disentangle primary dependencies from induced ones. 
Furthermore,  noise levels are not directly extractable. On the other hand 
its computational effort  scales favorable  with the number of data points as $N\log N$ 
as compared to $N^2$ with the proposed method. Our approach has its roots 
in autocorrelation methods \cite{grass,brock,savit}, but is conceptionally 
very distinct from these since it is based on continuity.  For that reason,  
in contrast to refs. \cite{grass,brock,savit}, it extracts the noise levels and 
ignores induced dependencies.

{\bf Acknowledgements}. This work was supported in part by the the Swedish Board for Industrial and 
Technical Development  (NUTEK) and the G\"{o}ran Gustafsson 
Foundation for Research in Natural Science and Medicine. One of the authors 
(Pi) gratefully acknowledges the funding provided by the Advanced
Research Projects Agency and the Office of Naval Research under grant 
ONR N000-14-92-J-4062.  Pi would also like to thank Brian D. Ripley
for helpful discussions.

\end{document}